\documentclass{article}

     \PassOptionsToPackage{numbers, compress}{natbib}
     \usepackage[preprint]{neurips_2023}

\usepackage[utf8]{inputenc} % allow utf-8 input
\usepackage[T1]{fontenc}    % use 8-bit T1 fonts
\usepackage{hyperref}       % hyperlinks
\usepackage{url}            % simple URL typesetting
\usepackage{booktabs}       % professional-quality tables
\usepackage{amsfonts}       % blackboard math symbols
\usepackage{nicefrac}       % compact symbols for 1/2, etc.
\usepackage{microtype}      % microtypography
\usepackage{xcolor}         % colors
\usepackage{amsmath}
\usepackage{subcaption}
\usepackage{graphicx}
\setcitestyle{square}

\title{Confidence Calibration for Systems with Cascaded Predictive Modules}

\author{Yunye Gong, Yi Yao, Xiao Lin, Ajay Divakaran, Melinda Gervasio\\
SRI International\\
\texttt{first.last@sri.com}}

\begin{document}

\maketitle

\begin{abstract}
  %The abstract paragraph should be indented \nicefrac{1}{2}~inch (3~picas) on
  %both the left- and right-hand margins. Use 10~point type, with a vertical
  %spacing (leading) of 11~points.  The word \textbf{Abstract} must be centered,
  %bold, and in point size 12. Two line spaces precede the abstract. The abstract
  %must be limited to one paragraph.
  Existing conformal prediction algorithms estimate prediction intervals at target confidence levels to characterize the performance of a regression model on new test samples. However, considering an autonomous system consisting of multiple modules, prediction intervals constructed for individual modules fall short of accommodating uncertainty propagation over different modules and thus cannot provide reliable predictions on system behavior. We address this limitation and present novel solutions based on conformal prediction to provide prediction intervals calibrated for a predictive system consisting of cascaded modules (e.g., an upstream feature extraction module and a downstream regression module). Our key idea is to leverage module-level validation data to characterize the system-level error distribution without direct access to end-to-end validation data. We provide theoretical justification and empirical experimental results to demonstrate the effectiveness of proposed solutions. In comparison to prediction intervals calibrated for individual modules, our solutions generate improved intervals with more accurate performance guarantees for system predictions, which are demonstrated on both synthetic systems and real-world systems performing overlap prediction for indoor navigation using the Matterport3D dataset.
\end{abstract}

\section{Introduction}
Despite recent successes in modeling real-world systems via data-driven machine learning technologies (e.g., deep neural networks) for achieving high accuracy which rivals, or even surpasses, human capabilities in various predictive tasks, one of the major concerns remained for state-of-the-art machine learning solutions is their limited capabilities in providing reliable performance guarantees and quantifying uncertainties of predictions~\cite{safe}. Since the lack of assured solutions prohibits confident application of real-world autonomous systems, especially safety-critical systems for tasks such as medical diagnosis and autonomous driving, proper calibration technologies are desired to reliably indicate how much a user should trust system predictions for decision-making~\cite{health,driving}.\par
Classic techniques address this problem by generating confidence scores for classification predictions~\cite{platt99,guo17} or prediction intervals for regression predictions~\cite{cp,scp}, to enable reliable prediction of model behavior given new test samples. In this work, we focus on the later setting for confidence prediction on system with continuous outputs. Specifically, conformal prediction algorithms~\cite{cp,scp,wcp} provide prediction intervals containing prediction ground-truths at given probabilities. The classic approach of split conformal prediction~\cite{scp} computes residues of regression predictions on a set of hold-out validation data and use empirical quantiles of residues to adjust new predictions. This approach is agnostic to data distributions and regression algorithms. More recent approaches extend the method to generate proper prediction intervals considering data heteroscedasticity~\cite{cqr} or distribution shifts such as the covariate shift~\cite{wcp}. While existing algorithms focus on individual predictive modules, real-world applications often involve large autonomous systems consisting of multiple modules which can be trained and updated independently~\cite{sys3,sys4,sys1,sys2}. For large autonomous systems, it is frequently challenging to acquire system-level data (or correspondence between module-level data) for end-to-end calibration due to limited resources, and security or privacy concerns. On the other hand, modules calibrated independently are not guaranteed to provide reliable confidence measures for system-level predictions. 
To demonstrate the unique challenge of system-level calibration, we start with a system with cascaded predictive modules (e.g., an upstream feature extraction module and a downstream regression module). In this setting, having an imperfect upstream module implies a shift in the input distribution for the downstream module from its ideal training or calibration distribution to the test distribution corresponding to erroneous upstream predictions. Furthermore, incorrect upstream predictions also lead to discrepancies between correct downstream predictions and correct system-level predictions. Consequently, even perfect regression predictions and performance guarantees calibrated for a downstream module can still be inaccurate for the overall system.\par
Motivated by the aforementioned challenges, we propose novel solutions to construct proper prediction intervals for system predictions addressing module-level interactions without end-to-end system-level data. Our key idea extends conformal prediction algorithms where prediction intervals are estimated based on quantiles of the empirical distribution of prediction errors at target coverage probabilities. While empirical quantiles for the system-level error distribution cannot be directly computed without system-level validation data, we instead leverage module-level validation data to characterize how module-level errors affect the system behavior. We construct solutions estimating an upper bound of system-level error and corresponding empirical quantiles for construction of safe prediction intervals. Inspired by recent development in similarity-based calibration~\cite{kiri, iccv21}, we further propose an improved solution for generating less conservative intervals better matching target coverage probabilities. We cluster module-level validation data based on similarity defined with respect to the Euclidean distance between upstream outputs (equivalently downstream inputs) and construct prediction intervals calibrated with respect to local error distributions. As shown in Sec.~\ref{sec:results}, we compare proposed solutions with two sets of baselines. We first apply split conformal prediction~\cite{scp} to the end-to-end system to show the target performance assuming ideal availability of system-level validation data. We also demonstrate that direct application of module-level conformal prediction to the downstream module cannot faithfully reflect the uncertainty in system-level predictions, even with the algorithms accommodating input distribution shifts introduced by imperfect upstream predictions~\cite{wcp,aci}. In comparison, our proposed solutions generate safer prediction intervals which contain system-level test ground-truths at probabilities better matching the targets. 

Our major contributions include the following:

(1) We identify an unexplored yet critical problem of providing reliable performance guarantees for system with cascaded modules. We propose novel calibration solutions constructing proper prediction intervals which allow system users to perform safe and informative predictions of system behavior without needing system-level validation data. 

(2) Our proposed solutions exploit the idea of similarity-based calibration with small sets of module-level validation data. We provide theoretical justification of proposed solutions based on the estimation of upper bounds for the target system-level error and empirical quantiles.

(3) We justify the proposed solutions with experimental results using both synthetic and real-world data. We demonstrate that module-level algorithms are not sufficient for system-level uncertainty characterization, while proposed system-level solutions provide improved prediction intervals with empirical coverage rates better matching the targets consistently over different target levels.

\section{Related Work}
\noindent\textbf{Model Calibration}. Model calibration seeks to generate accurate characterization of model competency for reliable prediction of model performance given new test samples, often via post-hoc calculations based on small sets of validation data. For deep classification models, popular approaches include temperature scaling, and its extension to vector scaling and matrix scaling~\cite{guo17}. Those methods learn additional calibration parameters given logits to ensure that the confidence score (e.g., maximum softmax probability) better matches the true probability of correct classification. For regression models with continuos outputs, conformal prediction approaches~\cite{cp,scp} generate, for each new test sample, a prediction interval which contains the ground-truth observation at a target confidence level. Assuming exchangeability between training and testing samples, classic split conformal prediction~\cite{scp,papadopoulos} provides an efficient and general solution which is not subject to a specific data distribution and generates intervals with guaranteed coverage rate by estimating empirical quantiles from hold-out validation data. Marx et al.~\cite{mcc} introduce a general framework which transforms various types of regression models to calibrated probabilistic models and unifies classes of algorithms including conformal prediction and conformal calibration~\cite{vovk20}.\par
\noindent\textbf{Calibration given distributional shifts}. More recent efforts address the challenge of model calibration in the context of transfer learning. For deep classification models, solutions extending temperature scaling are developed for unsupervised domain adaptation~\cite{wang20,pampari20} and domain generalization~\cite{iccv21} under the assumption of covariate shift. For regression models, Tibshirani et al. propose Weighted Conformal Prediction~\cite{wcp} which addresses covariate shift by estimating weighted empirical quantiles on validation errors from the source distribution. The weights can be estimated based on the density ratio between source and target data using unsupervised data. Podkopaev and Ramdas~\cite{podkopaev} address the problems of conformal prediction and classification calibration under the assumption of label shift. Gibbs and Cand\`es propose adaptive conformal prediction algorithms~\cite{aci,faci} which gradually adjust calibration parameters in an online learning setting to address arbitrary distributional shifts in data and achieve target coverage rate over long time intervals.\par
 However, these approaches do not specifically address the challenge of uncertainty quantification in multi-module systems. In comparison, we propose novel solutions directly exploiting module-level interactions to generate prediction intervals covering system-level test observations at desired probabilities, without using end-to-end system-level data.\par
\noindent\textbf{Similarity-based calibration.}
Several recent approaches improve model calibration by considering local heterogeneity within data. Such heterogeneity can affect the performance of global calibration solutions even without domain-level distribution shifts introduced at test time. For image classification models, calibration solutions have been proposed with calibration parameters specific to each class~\cite{class}, image domain~\cite{yu22}, cluster in the feature space~\cite{iccv21} or sample instance~\cite{kiri}. In the context of conformal prediction for regression models, global methods such as the split conformal prediction use the same empirical quantile of validation errors to correct all future predictions and thus generate prediction intervals with a fixed width. Romano et al.~\cite{cqr} extend the classic approach by incorporating quantile regression to generate intervals with widths varying based on the data to achieve potentially narrower prediction intervals while maintaining the coverage guarantee. Our proposed method is inspired by efforts exploiting similarity in image features and optimizes local calibration parameters.
\vspace{-1em}
\section{Formulation}\label{sec:formulation}
%\subsection{Predictive System with two cascaded modules}
\vspace{-1em}
We demonstrate the calibration of a multi-module system considering two cascaded predictive modules including an upstream predictive (e.g., detection, recognition) module and a downstream regression module with a continuous output variable (e.g., the reward of a reinforcement learning system). Let \( f: \mathbb{R}^{m} \rightarrow \mathbb{R}^{l}\) and \(g: \mathbb{R}^{l} \rightarrow \mathbb{R}\) denote the upstream and downstream oracle processes respectively. Let \(X\in \mathbb{R}^{m}, Y\in\mathbb{R}^{l}, Z\in\mathbb{R}\) denote the input, intermediate and output variables in the system. Let \(\hat{f}\) and \(\hat{g}\) denote the upstream and downstream modules learned to approximate \(f\) and \(g\) respectively. Motivated by real-world development of large systems, we assume different modules are developed in parallel and trained independently with data from ideal data distributions \(P_{X,Y}\) and \(P_{Y,Z}\). Assuming the availability of small sets of validation data from the same training distribution, existing solutions can provide accurate characterization of module outputs assuming new test samples from the same distribution. However, when the modules are composed into an end-to-end system at test time, performance predictions made based on module-level calibration can be inaccurate considering distribution shifts introduced by system composition. In particular, the input distribution for the downstream module \(\hat{g}\) at test time will be determined by imperfect upstream predictions from the learned upstream module \(\hat{f}\) deviating from its ideal training distribution. The scale of shifts corresponds to the level of accuracy of upstream predictions. Furthermore, imperfect upstream predictions also imply discrepancies between downstream supervision and system-level supervision leading to further shifts in the error distribution. To enable reliable competency prediction of the system, we propose system-level calibration algorithms to accommodate the distribution shifts implied, given alternative assumptions on data availability.

\subsection{End-to-end system-level calibration}\label{sec:end-to-end}
First, we consider a baseline solution assuming at least a small set of data with system-level supervision is available for the purpose of calibration. While this assumption provides ideal calibration data to generate proper competency predictions, it is less realistic as additional system-level data can be either expensive or fundamentally challenging to collect in real-world development, especially for users operating the system in a test mode. We describe this setting as it serves a baseline establishing the target performance level for system-level calibration. Specifically, given a set of validation data drawn from the joint distribution \(P_{X,Z}\), the end-to-end system \((\hat{g}\circ\hat{f})(\cdot)\) can be considered as one black-box regression function. With no distribution shifts assumed at system-level, the problem is reduced as the interaction between modules with respect to error and confidence propagation can be ignored. In this setting, we adopt the split conformal prediction ~\cite{papadopoulos,scp} for generation of prediction intervals with performance guarantees considering an end-to-end regression system.

Given an end-to-end regression system denoted as \(\hat{\mu}=\hat{g}\circ\hat{f}\) and a set of system-level validation data \(D=\{ (X_i,Z_i) \}_{i=1}^n\) assumed to be independently drawn from the distribution \(P_{X,Z}\), we seek to construct a proper prediction interval \(C(X_{n+1})\) given a new test sample \(X_{n+1}\) such that we can be confident about the presence of the corresponding system-level ground-truth \(Z_{n+1}\) in the interval.
Following~\cite{papadopoulos, scp}, we compute the system-level validation error, denoted as \(S_i=|Z_i-\hat{\mu}(X_i)|\), for each \((X_i,Z_i)\in D\). We compute the empirical quantile of the prediction error \(S\) on the validation data set \(D\) given a target coverage probability \(\alpha\) and denote it as \(Q_{\alpha,D}(S)\). Specifically, assuming system-level validation errors \(S_1,...,S_n\) are exchangeable samples with order statistics \(S_{(1)}\leq...\leq S_{(n)}\), the quantile at probability \(\alpha\) with respect to the empirical distribution is computed as
\begin{align}
    Q_{\alpha,D} (S) = 
     \begin{cases}
      S_{(\lceil(n+1)\alpha\rceil)} &\text{if} \lceil(n+1)\alpha\rceil \leq n \\
      \infty & \text{otherwise},
     \end{cases}
\end{align}
Following key lemmas on quantiles utilized in conformal prediction approaches~\cite{scp,cqr,wcp}, assuming \(S_1\),...\(S_{n+1}\) are exchangeable, for any \(\alpha\in(0,1)\), we have
\begin{align}
\mathbb{P}\{S_{n+1}\leq Q_{\alpha,D}(S)\}\geq \alpha.
\label{eq:quantile_lemma1}
\end{align} If we can further assume $S_1,...,S_{n+1}$ to be almost surely distinct, then we have
\begin{align}
    \mathbb{P}\{S_{n+1}\leq Q_{\alpha,D}(S)\}\leq \alpha+\frac{1}{n+1}.
\label{eq:quantile_lemma2}
\end{align}
Given a test sample $X_{n+1}$, we construct the prediction interval as 
\begin{align}
C(X_{n+1})=[\hat\mu(X_{n+1})-Q_{\alpha,D}(S),\hat\mu(X_{n+1})+Q_{\alpha,D}(S)].\label{eq:qs} 
\end{align}
Following Eq.~\ref{eq:quantile_lemma1} and \ref{eq:quantile_lemma2},
the coverage guarantee for constructed prediction intervals~\cite{scp,wcp} is given as
\begin{align}
    \mathbb{P}\{Z_{n+1}\in C(X_{n+1})\}\geq \alpha.
\end{align} assuming $(X_i,Y_i)$ for $1\leq i\leq n+1$ are exchangeable. Furthermore, if the errors are almost surely distinct, then the probability is also bounded above as
\begin{align}
    \mathbb{P}\{Z_{n+1}\in C(X_{n+1})\}\leq \alpha + \frac{1}{n+1}.
\end{align}
\vspace{-1.5em}
\subsection{System-level calibration with module-level data}
In this section, we propose system-level calibration without system-level data to address the challenge of data scarcity in real-world system applications. Given only module-level data from training distributions of individual modules, we propose algorithms to estimate the test distribution of system-level prediction error $S$. Let $D_f=\{(X_i,Y_i)|i\in\mathcal{I}_f\}$ denote a set of validation data for the upstream module, drawn from its ideal training distribution $P_{X,Y}$, and let $D_g=\{(Y_i,Z_i)|i\in\mathcal{I}_g\}$ denote a set of validation data for the downstream module, drawn from its ideal training distribution $P_{Y,Z}$, where $\mathcal{I}_f$ and $\mathcal{I}_g$ refer to two sets of random indices that are not guaranteed to overlap. In comparison to the end-to-end system-level calibration solution specified in Sec.~\ref{sec:end-to-end}, we target proper characterization of system behavior without pairwise correspondence between data in $D_f$ and $D_g$. We refer to this method as set-level calibration with module-level data.\par
Specifically, based on upstream validation data $D_f$, we estimate the empirical distribution of upstream prediction errors propagated through the downstream predictor $\hat{g}$ by computing
\begin{align}
    U_i = |(\hat{g}\circ\hat{f})(X_i) - (\hat{g}\circ f)(X_i)| = |\hat{g}(\hat{f}(X_i)) - \hat{g}(Y_i)|\label{eq:uperr}
\end{align} for each $(X_i,Y_i)$ pair in $D_f$. Based on downstream validation data, we estimate the empirical distribution of downstream prediction errors assuming perfect upstream predictions by computing
\begin{align}
    W_j = |(\hat{g}\circ f)(X_j) - (g\circ f)(X_j)|=|\hat{g}(Y_j)-Z_j|\label{eq:downerr}
\end{align} for each $(Y_j,Z_j)$ pair in $D_g$. Following Triangle Inequality~\cite{TI}, we have \begin{align}
    S_i &= |(\hat{g}\circ\hat{f})(X_i) - (g\circ f)(X_i)|
     = |(\hat{g}\circ\hat{f})(X_i)- (\hat{g}\circ f)(X_i) + (\hat{g}\circ f)(X_i) - (g\circ f)(X_i)|
     \nonumber\\
     &\leq |(\hat{g}\circ\hat{f})(X_i) - (\hat{g}\circ f)(X_i)|+|(\hat{g}\circ f)(X_i) - (g\circ f)(X_i)|= U_i+W_i,
\end{align} which suggests that the two error scores defined in Eq.~\ref{eq:uperr} and Eq.~\ref{eq:downerr} can be used to compute an upper bound for the target system-level error $S$ when the two sets of data samples are aligned. In an ideal scenario as specified in Sec.~\ref{sec:end-to-end}, we compute the empirical quantile of system-level prediction errors $S$ on validation data at a given probability. However, without the correspondence between module-level validation data, we cannot directly compute empirical quantiles of the target end-to-end errors. Instead, we compute empirical quantiles separately for module-level errors $U$ and $W$ and estimate an upper bound of the quantiles.\par
Consider $A$ and $B$ as two scalar random variables, an upper bound for the quantile of the sum of two variables $A+B$ at probability $\alpha\in(0,1)$ is given as: 
\begin{align}
    Q_{\alpha}(A+B)\leq\min_{\beta\in[\alpha,1]}[Q_{\beta}(A)+Q_{1-\beta+\alpha}(B)]
\end{align} We refer to~\cite{qsum} for the detailed proof of this given bound. Based on module-level data, we compute
\begin{align}
    \hat{Q}_{\alpha,D_f,D_g} := \min_{\beta\in[\alpha,1]}[Q_{\beta,D_f}(U)+Q_{1-\beta+\alpha,D_g}(W)]\geq Q_{\alpha,D}(S).\label{eq:qm}
\end{align} %which serves an estimate of $Q_{\alpha,D}(S)$. 
Then following Eq.~\ref{eq:qs}, given a new test sample $X_t$, we construct a safe yet conservative interval as
\begin{align}
    \hat{C}_{D_f,D_g}(X_t)=[&(\hat{g}\circ\hat{f})(X_t)- \hat{Q}_{\alpha,D_f,D_g},(\hat{g}\circ\hat{f)}(X_t)+ \hat{Q}_{\alpha,D_f,D_g}].
\end{align} In comparison to the intervals computed based on end-to-end data, the intervals approximated with only module-level data target a given coverage probability subject to wider interval widths as
%\begin{align}
   \(\mathbb{P}\{Z_{t}\in \hat{C}_{D_f,D_g}(X_t)\} \geq \mathbb{P}\{Z_{t}\in C(X_{t})\}\geq \alpha\).
%\end{align}. 
\subsection{System-level calibration with clustering}
While $\hat{Q}_{\alpha,D_f,D_g}$ as defined in Eq.~\ref{eq:qm} serves an upper-bound estimate of the target quantile $Q_{\alpha,D}$ of end-to-end system level validation errors, it is approximated using set-level validation data and can be overly conservative as demonstrated in Sec.~\ref{sec:results}. To improve the system-level prediction intervals, we further propose an alternative algorithm exploiting the relative position of each sample within the set-level data distribution and thus relaxing the constraints of fixed interval widths for all samples. This method is inspired by approaches in confidence calibration for classification models where local optimal calibration parameters, instead of a global set of calibration parameters, are learned for improved confidence predictions~\cite{kiri}.\par
Given module-level validation data $D_f=\{(X_i,Y_i)|i\in\mathcal{I}_f\}$ and $D_g=\{(Y_i,Z_i)|i\in\mathcal{I}_g\}$ independently drawn from the ideal training distribution of the corresponding module, we assume no end-to-end supervision and thus no one-to-one correspondence available between the two sets of validation data. We propose to estimate local correspondence based on the Euclidean distance between samples in the space of intermediate variable $Y$. For each set of module-level validation data, We perform $K$-means clustering~\cite{kmeans} based on $Y$ values. We denote the resulting clusters as $\{D_f^{c_i} \}_{i=1}^{Nc}$ and $\{ D_g^{c_i} \}_{i=1}^{Nc}$ for $D_f$ and $D_g$ respectively. We then form cluster-level correspondence for two sets of module-level validation data based on the Euclidean distance between cluster centroids. Given a clustering of upstream validation data, we consider a corresponding clustering for upstream predictions given $\hat{f}$. Given a new test sample $X_t$, we compute the upstream prediction $\hat{f}(X_t)$ and perform nearest neighbor prediction to identify a cluster of upstream validation data $D_f^{c_i}$ based on Euclidean distances between cluster centroids and the sample prediction in the space of $\hat{f}(X)$. Given $D_f^{c_i}$, we identify the cluster of downstream validation data $D_g^{c_j}$ with the nearest centroid.  We then estimate the empirical quantile parameters as specified in Eq.~\ref{eq:qm} at cluster-level and compute
\begin{align}
    \hat{Q}_{\alpha,D_f^{c_i},D_g^{c_j}} = \min_{\beta\in[\alpha,1]}[Q_{\beta,D_f^{c_i}}(U)+Q_{1-\beta+\alpha,D_g^{c_j}}(W)].\label{eq:qc}
\end{align}
We construct system-level prediction interval as
\begin{align}
    \hat{C}_{D_f^{c_i},D_g^{c_j}}(X_t)=[(\hat{g}\circ\hat{f})(X_t)- \hat{Q}_{\alpha,D_f^{c_i},D_g^{c_j}}, (\hat{g}\circ\hat{f)}(X_t)+ \hat{Q}_{\alpha,D_f^{c_i},D_g^{c_j}}].
\end{align}
\section{Experiments}
\subsection{Datasets}
\noindent\textbf{Simulated System.}
To demonstrate our proposed system-level calibration algorithms, we perform experiments on both simulated and real-world data. For a simulated system, we simulate two cascaded modules each as a random linear transformation~\cite{initial} performing $\mathbb{R}^{64}\rightarrow\mathbb{R}^{32}$ and $\mathbb{R}^{32}\rightarrow\mathbb{R}$ transformations respectively. 
We simulate the system (i.e., upstream) inputs as random samples from a unit Gaussian distribution.
To demonstrate the effect of having an imperfect upstream module on system competence prediction, we introduce addictive Gaussian noises with varying statistics to upstream predictions. The effect of upstream prediction noise is shown in Fig.~\ref{fig:noise_var_and_mean}. We perform Random Forest regression for learning the downstream module. We randomly split simulated data into disjoint sets for downstream regression training, system-level validation, upstream module validation, downstream module validation and testing. To demonstrate the effect of data size on proposed algorithms, we perform multiple experiments using 500 to 8000 data samples for training and calibration. For each experiment, we use 5000 holdout samples for testing. For experiments with 500 samples, we repeat experiments with 500 trials over different random splits of data to compute the standard deviation of the empirical coverage, as shown in Fig.~\ref{fig:sim_and_matterport} (a). Baseline comparison over 500 trials takes around 6.5 CPU hours. 

\noindent\textbf{Matterport3D.}
To further demonstrate the proposed algorithms on a real-world system, we consider a system for indoor scene understanding including an upstream depth estimation module and a downstream view overlap prediction module based on Matterport3D dataset~\cite{matterport}. The dataset contains 194,400 RGB-D images of indoor scenes from 90 buildings. We select this dataset for sufficient complexity to support cascaded system  with ground-truth for evaluation. We access the dataset with consent under non-commercial license. We use provided ground-truth depth maps and ground-truth overlap ratios given pairs of scenes for our experiments. For experiments reported in Sec.~\ref{sec:results}, we use data from a random subset of 10 different buildings. We use RGB images as system inputs and apply a pretrained monocular depth estimator~\cite{LeReS} with ResNet50 backbone~\cite{resnet} to generate upstream predictions. We rescale each depth estimation to 125x100 dimension and perform downstream regression with a multilayer perceptron (MLP) which is trained to predict a scalar overlap ratio between two corresponding scenes given a pair of depth maps. We randomly split the data into disjoint subsets and use 500 samples for training the downstream regression module, 500 samples for estimating the calibration parameters and use 5000 samples for testing. The experiments with larger sets of training and calibration data are included in the supplementary materials.

\subsection{Experimental settings}
\noindent\textbf{Baseline: module-level calibration.}
To demonstrate the necessity of system-level calibration algorithms, we first perform baseline experiments applying state-of-the-art conformal correction algorithms addressing only the downstream module in a system with two cascaded modules. We consider two baseline approaches including Weighted Conformal Prediction (WCP)~\cite{wcp} which generalizes conformal prediction to address input distribution shifts from training to testing and Adaptive Conformal Inference (ACI)~\cite{aci} which targets proper coverage given varying marginal data distributions. We emphasize that, while these methods are designed to compute accurate prediction intervals for module-level predictions, they do not specifically address the challenging scenario of system-level calibration given no data samples from the target (end-to-end) distribution available.
For each set of experiments with either simulated or real-world system, We use samples randomly drawn from the ground-truth distribution $P_{Y,Z}$ to train the downstream module. For WCP, we follow the algorithm reported in original paper~\cite{wcp} and use unsupervised data from the source (downstream training) and the target (upstream prediction) distributions to estimate the density ratio which is used to correct the quantile estimation. For ACI~\cite{aci}, with no supervised data from end-to-end test distribution available at calibration stage, we adapt the conformal inference to module-level data distribution. We use system-level data drawn from $P_{X,Z}$ only for evaluation. As shown in Sec.~\ref{sec:results}, while these algorithm accommodates marginal distribution shifts for the downstream module, it is still insufficient for generating prediction intervals containing system-level ground-truths with confidence. Given an inaccurate upstream prediction, even perfectly accurate downstream prediction can be far away from the system target. Therefore, prediction intervals calibrated with respect to module-level ground-truths are not directly applicable for reliable characterization of system behavior. \par
\noindent\textbf{System-level calibration with end-to-end data.}
To set the target performance of system-level calibration, we perform baseline experiments using ideal end-to-end validation data. We sample downstream data from $P_{Y,Z}$ for training the individual module. We use a set of end-to-end data randomly drawn from $P_{X,Z}$ for computing conformal correction. We evaluate the generated prediction intervals on a disjoint set of data drawn from $P_{X,Z}$.\par
\noindent\textbf{System-level calibration with module-level data.}
For system-level calibration experiments using only module-level validation data, we use data randomly drawn from $P_{Y,Z}$ to train the downstream module. We use disjoint sets of validation data drawn from $P_{Y,Z}$ and $P_{X,Y}$ to estimate the upper bounds of the target end-to-end system-level error distribution and corresponding quantile parameters. For system-level calibration with clustering, we perform $K$-means clustering~\cite{kmeans} on each set of module-level validation data, based on the Euclidean distance in the space of intermediate variable $Y$. The effect of varying numbers of clusters is studied and is shown in Fig.~\ref{fig:cluster}. We use different numbers of clusters for experiments with different sizes of validation data, targeting 10 samples per cluster on average. We evaluate the generated prediction intervals on a disjoint set of data drawn from \(P_{X,Z}\).
\begin{table*}[!t]
\caption{Empirical coverage (\%) and average width of prediction intervals generated for simulated system. Performance averaged over 500 trials of random data splits. For each trial, disjoint sets of data are used for training (500 samples), calibration (500 samples) and testing (5000 samples). Upstream prediction errors are simulated from a unit Gaussian. 
%\vspace{-1em}
}
\label{tab:sim}
\centering
\small
\setlength\tabcolsep{4pt}
%\vspace{3pt}
\begin{tabular}{rcccccccccc}
\hline
&\multicolumn{2}{c}{\shortstack{\\WCP~\cite{wcp} on\\Downstream}}
&\multicolumn{2}{c}{\shortstack{ACI~\cite{aci} on \\Downstream}}
&\multicolumn{2}{c}{End-to-end} &\multicolumn{2}{c}{\shortstack{Set-level\\(ours)}} &\multicolumn{2}{c}{\shortstack{Cluster-level\\(ours)}}\\
\hline
%\backslashbox{method}{target}
\shortstack{\\Target}&\shortstack{Coverage}
&\shortstack{Width}&
\shortstack{Coverage}
&\shortstack{Width}&
\shortstack{Coverage}
&\shortstack{Width}&
\shortstack{Coverage}
&\shortstack{Width}&
\shortstack{Coverage}
&\shortstack{Width}\\
\hline
50\% & 38.02\% & 0.34 & 33.44\% & 0.30 & 49.99\% & 0.45 & 85.83\% & 0.99 & 79.30\% & 0.88\\
60\% & 46.05\% & 0.42 & 40.96\% & 0.37 & 59.97\% & 0.57 & 89.92\% & 1.11 & 83.53\% & 0.97\\
70\% & 54.21\% & 0.51 & 49.20\% & 0.46 & 69.91\% & 0.70 & 93.54\% & 1.25 & 87.49\% & 1.08\\
80\% & 63.34\% & 0.62 & 58.74\% & 0.57 & 79.78\% & 0.86 & 96.54\% & 1.42 & 91.15\% & 1.21\\
90\% & 74.62\% & 0.78 & 71.08\% & 0.74 & 89.80\% & 1.10 & 99.71\% & 2.01 & 94.50\% & 1.37\\
\hline
\end{tabular}
%\vspace{-1em}
\end{table*}

\begin{table*}[!]
%\vspace{-1em}
\caption{Empirical coverage (\%) and average width of prediction intervals generated on Matterport3D~\cite{matterport} systems. Disjoint sets of data are used for training (500 samples), calibration (500 samples) and testing (5000 samples). 
%\vspace{-1em}
}
\label{tab:matterport}
\centering
\small
%\vspace{3pt}
\setlength\tabcolsep{4pt}
\begin{tabular}{rrrrrrrrrrr}
\hline
&\multicolumn{2}{c}{\shortstack{\\WCP~\cite{wcp} on\\Downstream}}
&\multicolumn{2}{c}{\shortstack{ACI~\cite{aci} on \\Downstream}}
&\multicolumn{2}{c}{End-to-end} &\multicolumn{2}{c}{\shortstack{Set-level\\(ours)}} &\multicolumn{2}{c}{\shortstack{Cluster-level\\(ours)}}\\
\hline
%\backslashbox{method}{target}
\shortstack{\\Target}&\shortstack{Coverage} &\shortstack{Width}&
\shortstack{Coverage} &\shortstack{Width}&
\shortstack{Coverage} &\shortstack{Width}&
\shortstack{Coverage} &\shortstack{Width}&
\shortstack{Coverage} &\shortstack{Width}\\
\hline
50\% & 5.20\% & 0.11 & 3.00\% & 0.10 & 52.80\% & 0.54 & 71.32\% & 0.78 & 61.84\% & 0.68\\
60\% & 6.50\% & 0.13 & 5.08\% & 0.11 & 61.40\% & 0.62 & 74.68\% & 0.90 & 70.24\% & 0.77\\
70\% & 6.50\% & 0.13 & 8.00\% & 0.17& 71.44\% & 0.79 & 96.50\% & 1.21 & 77.76\% & 0.87\\
80\% & 14.98\% & 0.22 & 11.92\% & 0.21 & 79.44\% & 1.03 & 99.10\% & 1.34 & 84.34\% & 1.03\\
90\% & 27.50\% & 0.33 & 19.56\% & 0.30 & 90.94\% & 1.14 & 99.56\% & 1.50 & 90.78\% & 1.22\\
\hline
\end{tabular}
\vspace{-1em}
\end{table*}
\subsection{Results and Discussion}\label{sec:results}
We evaluate the performance of proposed algorithms with standard metrics~\cite{cqr,wcp} measuring the empirical coverage and the averaged widths of prediction intervals generated for system predictions.
The empirical coverage measures the percentage of prediction intervals that contain the prediction ground-truths. 
Having empirical coverage lower than the target implies over-confident predictions where the ground-truths tend to fall out of constructed intervals and poses higher risks in trusting the predictions. On the other hand, having empirical coverage higher than the target implies under-confident predictions and provides less information. For example, in an extreme case, trivial intervals with infinite lengths can always be generated to reach $100\%$ empirical coverage. With proposed algorithms, we target prediction intervals that can reach the target coverage rate with shorter widths.\par
\begin{figure*}[htp]
    \centering
    \begin{subfigure}{0.52\linewidth}
      \centering
      \includegraphics[width=\linewidth,trim=4 0 0 0,clip]{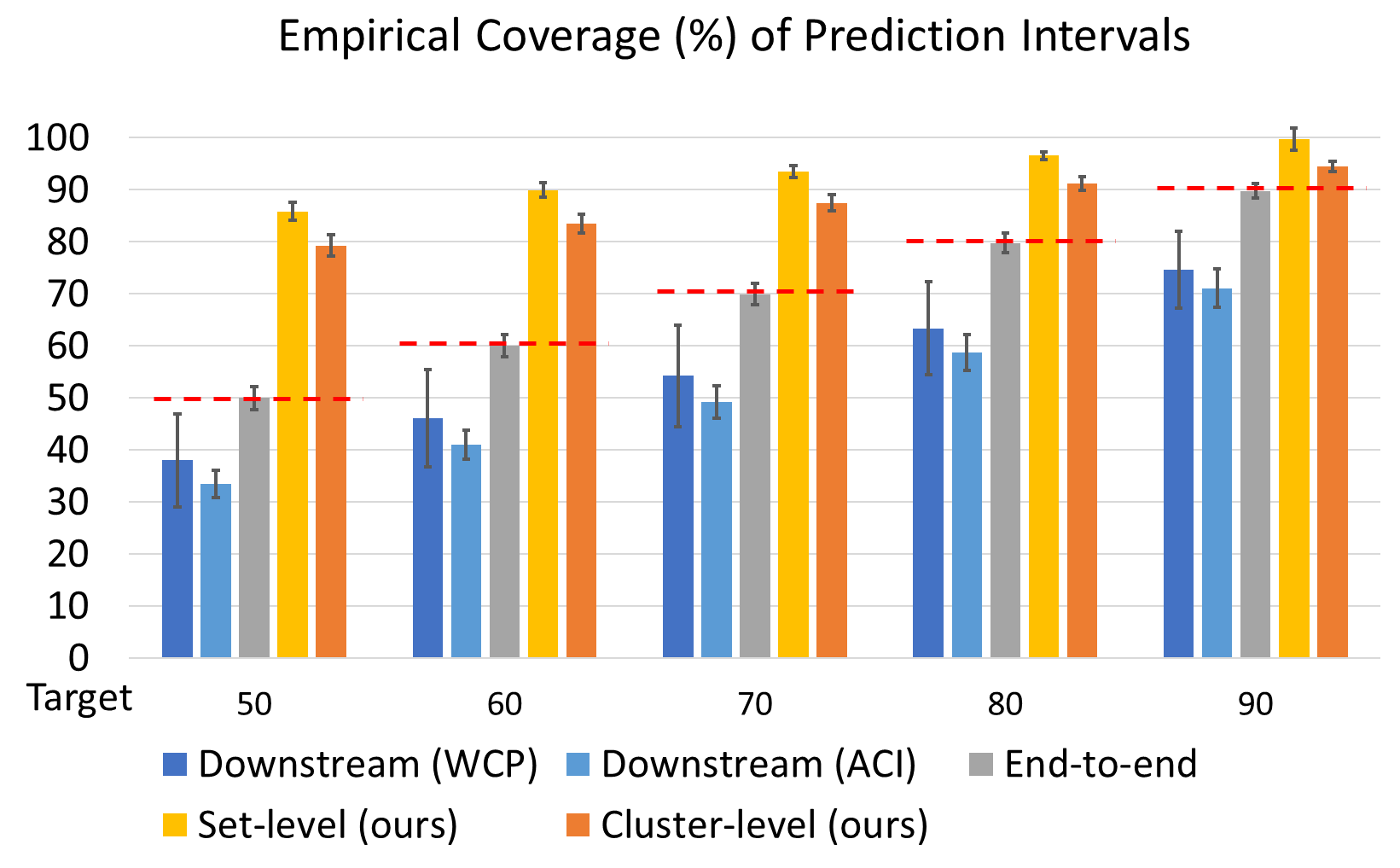}
      \caption{Simulated system.}
    \end{subfigure}%
    \begin{subfigure}{0.52\linewidth}
      \hspace{-1em}
      %\centering
      \includegraphics[width=\linewidth,trim=4 0 0 0,clip]{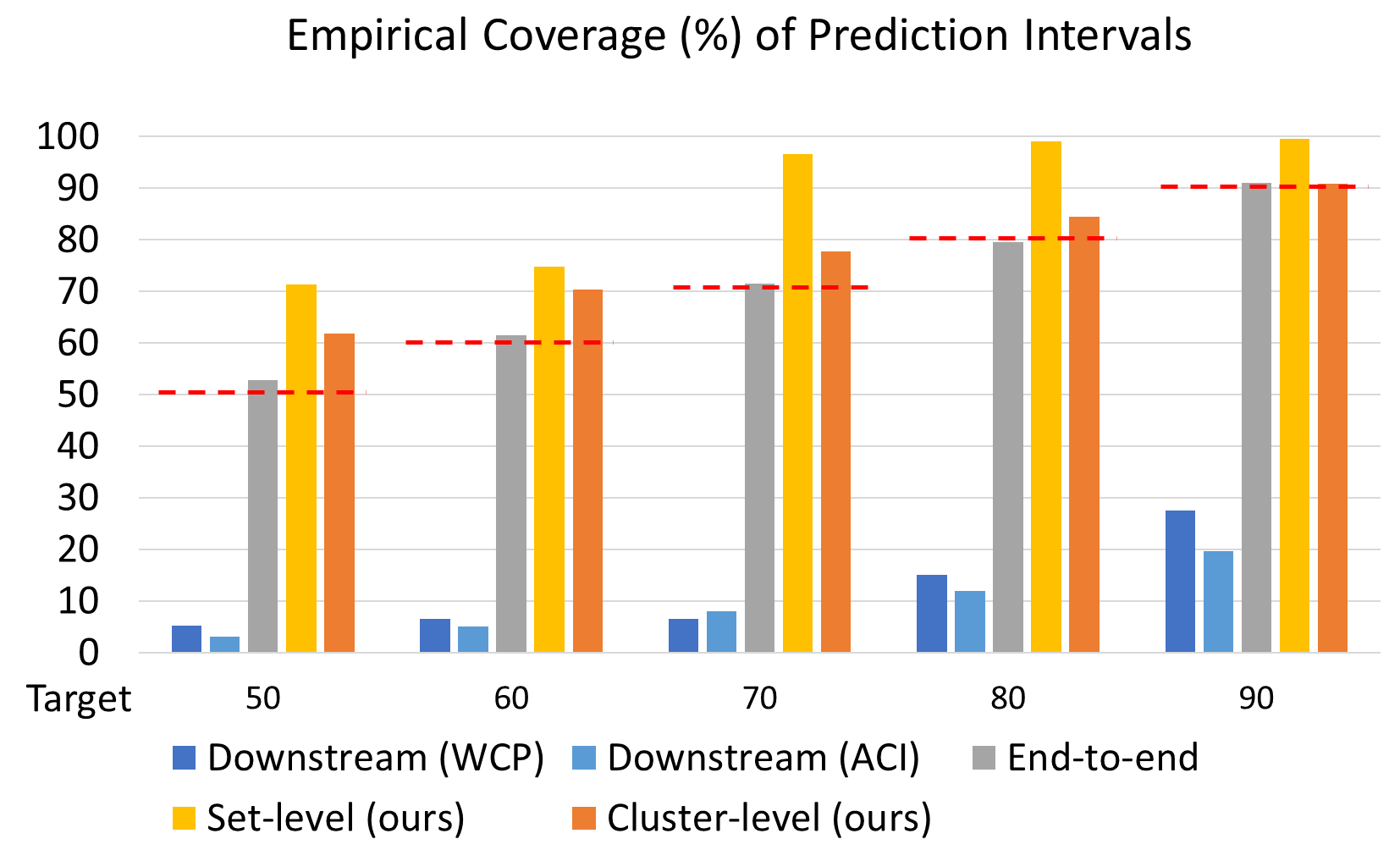}
      \caption{Matterport3D system.}
    \end{subfigure}
  \caption {Empirical coverage (\%) of prediction intervals on test data given different target coverage rates labeled in red. Error bars representing standard deviaton over 500 random data splits are included for experiments on simulated data and omitted for experiments on Matterport3D due to computational cost.}
\label{fig:sim_and_matterport}
\vspace{-1em}
\end{figure*}

\noindent\textbf{Comparison against downstream-only calibration.}
In leftmost panels in Table~\ref{tab:sim}-\ref{tab:matterport} (correspondingly dark and bright blue in Fig.~\ref{fig:sim_and_matterport}), we report the performance of module-level baselines~\cite{wcp,aci}. Results show that module-level calibration cannot provide reliable confidence quantification for system-level predictions. As an incorrect upstream prediction implies a mismatch between downstream target given downstream input and system target given system input, even well-constructed prediction intervals providing target coverage on module-level predictions will no longer be valid when used in a cascaded system. In fact, calibrating prediction intervals considering marginal data shifts on the downstream module results in largely under-coverage prediction intervals at system-level. In comparison, our proposed system-level calibration algorithms address the challenge by considering a set of system-level prediction errors that would be exchangeable to test samples from the target distribution and estimating the empirical quantile parameters corresponding to this ideal set of system errors either with or without end-to-end system-level data. In this way, we demonstrate largely improved prediction intervals mitigating the under-coverage problem of downstream-only baselines, in both simulated and real-world systems over a wide range of target coverage rates.\par

\begin{figure*}[htp]
    \centering
    \begin{subfigure}{0.48\linewidth}
      \centering
      \includegraphics[width=\linewidth]{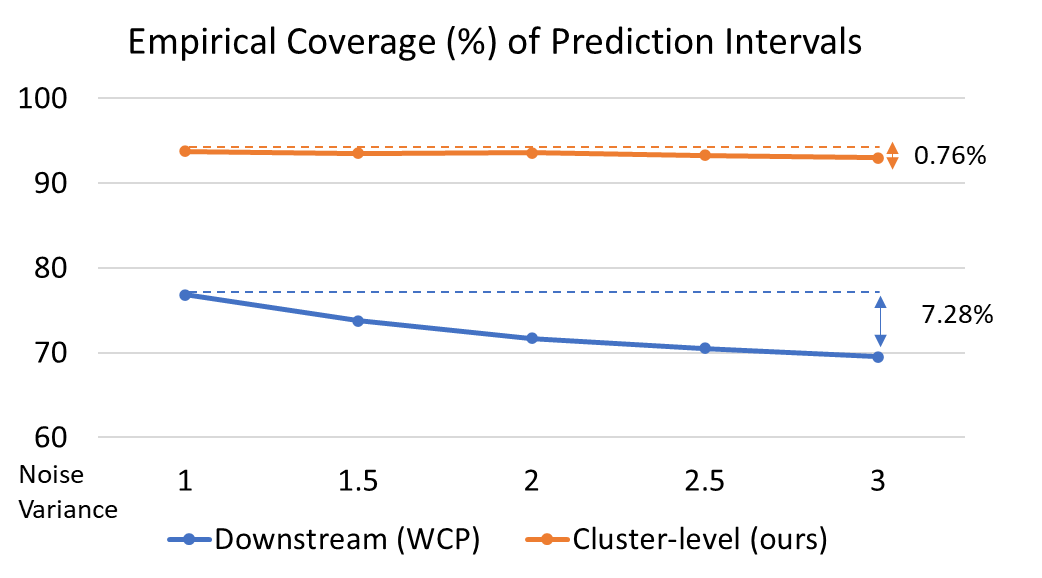}
      \caption{}
    \end{subfigure}%
    \begin{subfigure}{0.5\linewidth}
      \centering
      \includegraphics[width=\linewidth]{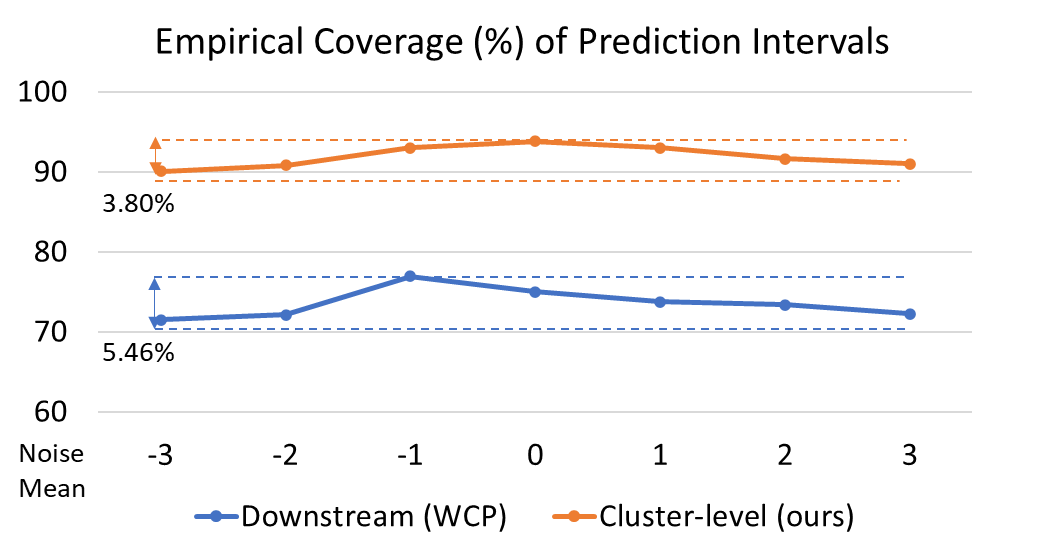}
      \caption{}
    \end{subfigure}
  \caption {Empirical coverage (\%) of prediction intervals given 90\% target coverage. Performance of simulated system with upstream prediction errors at different scales. Random noise samples drawn from (a) zero-mean Gaussian with varying variance and (b) unit-variance Gaussian with varying mean are added to upstream predictions.}
\label{fig:noise_var_and_mean}
\end{figure*}
\begin{figure*}[htp]
\vspace{-1em}
    \begin{minipage}{0.48\linewidth}
      \centering
      \includegraphics[width=\linewidth]{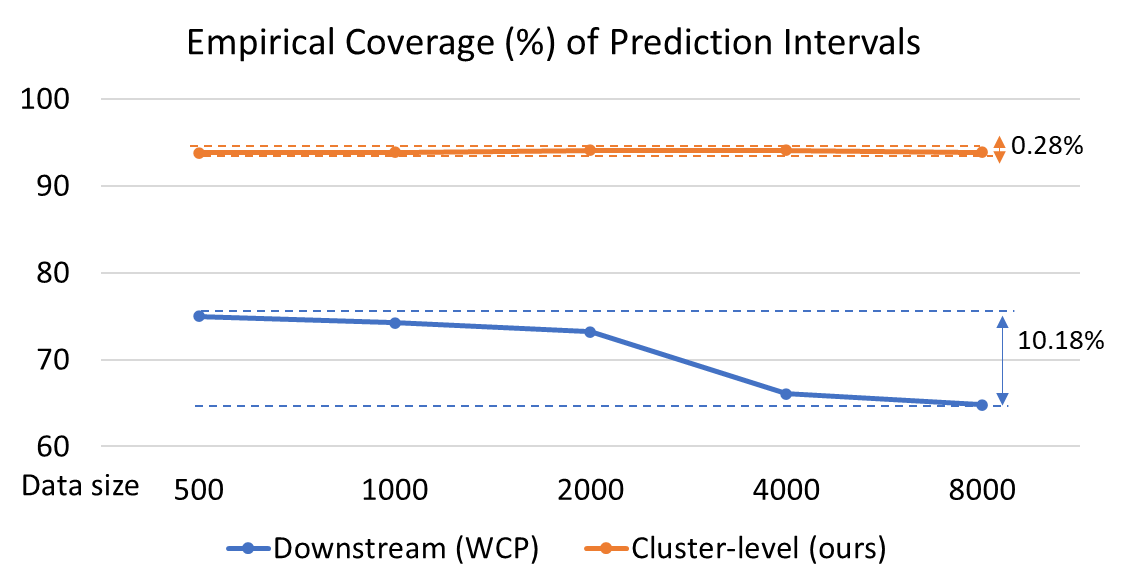}
      \caption{Empirical coverage (\%) of prediction intervals given 90\% target coverage. Performance of simulated system using different sizes of training and calibration data.}
      \label{fig:datasize}
    \end{minipage}\hfill
    \begin{minipage}{0.48\linewidth}
      \centering
      \includegraphics[width=\linewidth]{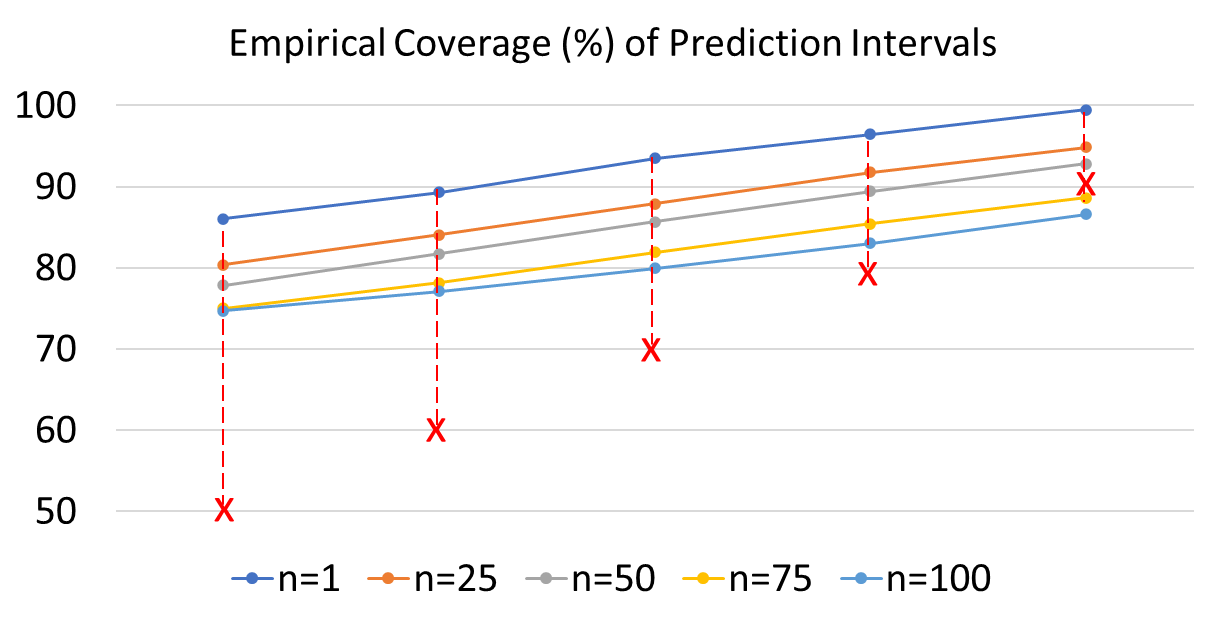}
      \caption{Performance of cluster-level calibration with 500 samples considering different number of clusters $n$. The target confidence levels are marked with red.}
      \label{fig:cluster}
    \end{minipage}
    \vspace{-1em}
\end{figure*}
\noindent\textbf{Comparison among system-level methods.}
When end-to-end data is available, the empirical quantile parameters of the target system-level errors can be directly computed. In this ideal setting, we demonstrate well-calibrated prediction intervals with empirical coverage closely matching the target at different levels, as shown in the ``End-to-end" Panel in Tabel~\ref{tab:sim}-\ref{tab:matterport} and correspondingly gray in Fig.~\ref{fig:sim_and_matterport}. In comparison, without direct access to end-to-end data, we propose solutions estimating an upper bound for the target quantile parameter, resulting in prediction intervals with empirical coverage rates consistently achieving the target rates at different levels, as shown in right two panels in Tabel~\ref{tab:sim}-\ref{tab:matterport}. In comparison to the baseline with ideal data, the prediction intervals generated with only module-level data are subject to wider intervals corresponding to safe yet relatively conservative predictions.
Comparing to the set-level solution using the same parameter for correction of all test predictions, the cluster-level solution exploits data heterogeneity and relaxes the implicit constraint on fixed interval width. Therefore, in comparison to set-level results (yellow in Fig.~\ref{fig:sim_and_matterport}), cluster-level results (orange in Fig.~\ref{fig:sim_and_matterport}) demonstrate consistently improved prediction intervals over different systems with reduced gaps between empirical and target coverage rates, corresponding to less conservative yet still safe prediction of system competency.\par
In Fig.~\ref{fig:noise_var_and_mean}, we demonstrate the robustness of proposed solution over different levels of accuracy of the upstream module. We perform experiments with simulated upstream predictions at different noise levels and demonstrate comparable performance. In fact, the proposed algorithms are generally applicable to modules with different prediction powers which correspond to different error distributions. 
In Fig.~\ref{fig:datasize}, we further demonstrate the robustness of proposed solutions over varying sizes of data used for training the regression models and calibrating the parameters for constructing prediction intervals. The empirical coverage rates of generated prediction intervals vary by only 0.28\% for our cluster-level solution, in comparison to the variation at 10.18\% for the baseline method considering data sizes ranging from 500 to 8000. Fig.~\ref{fig:cluster} describes the empirical performance of proposed cluster-level solution considering different numbers of clusters. We show that with an increasing number of clusters, we exploit data heterogeneity at an increasing granularity and achieve less conservative prediction intervals. While in experiments with $90\%$ target coverage, we also see that too many clusters can lead to under-coverage prediction intervals as not enough samples within each cluster are available for estimation of error distributions. For our experiments, we choose the number of clusters based on the size of validation data and target 10 samples per cluster on average to allow proper estimation of the empirical quantile parameters.
\vspace{-1em}
\section{Conclusion}
\vspace{-1em}
In this work, we address the problem of confidence calibration for predictive system with cascaded modules. Instead of improving the modules or the system with respect to prediction accuracy, we target informative and reliable characterization of system competency in the form of accurate prediction intervals for system predictions. We demonstrate that this problem is more challenging than calibration of individual (downstream) module given simple assumptions such as covariate shift, especially without access to additional end-to-end system-level data. Our key idea is to exploit module-level validation data to estimate an upper bound for system-level prediction errors and corresponding quantile parameters. We further leverage the idea of similarity-based calibration and estimate localized calibration parameters at cluster level. With experimental results on both simulated and real-world systems, we demonstrate that proposed system-level solutions can improve upon the under-coverage baselines by large margins and generate prediction intervals containing test ground-truths at high probabilities. The cluster-level solution further improves the resulting intervals, achieving empirical coverage rates better matching the targets. While current formulation only focuses on regression systems with two cascaded modules, the formulation can be generalized to multiple cascaded modules incrementally. It would be of interest for future research to explore uncertainty characterization considering larger and more complicated system configurations.

%\section{Acknowledgements}

%%%%%%%%% REFERENCES
{
\small
\bibliographystyle{nips}
\bibliography{egbib}
}

%acknowledgments

\end{document}